\documentclass[10pt,twocolumn,letterpaper]{article}

\usepackage{cvpr}              % To produce the CAMERA-READY version
\definecolor{cvprblue}{rgb}{0.21,0.49,0.74}
\usepackage[pagebackref,breaklinks,colorlinks,allcolors=cvprblue]{hyperref}
\usepackage{graphicx}
\usepackage{amsmath}
\usepackage{amssymb}
\usepackage{booktabs}
\usepackage{multirow}
\usepackage[accsupp]{axessibility}
\usepackage{tcolorbox}
\usepackage{pifont}
\usepackage{makecell}
\usepackage{xcolor}
\usepackage{colortbl}
\usepackage{stfloats}

\title{
DaMO: A Data-Efficient Multimodal Orchestrator for Temporal Reasoning with Video LLMs
}

\author{{{Bo-Cheng Chiu{$^{{1}}$}}\;\;
{Jen-Jee Chen{$^{{1}}$}}\;\;
{Yu-Chee Tseng{$^{{1}}$}}\;\;
{Feng-Chi Chen{$^{{2}}$}}\;\;
{An-Zi Yen{$^{{3}}$}}}
\\
{\small {College of Artificial Intelligence, National Yang Ming Chiao Tung University$^{1}$}}
\\
{\small {Institute of Population Health Sciences, National Health Research Institutes$^{2}$}}
\\
{\small {Department of Computer Science, National Yang Ming Chiao Tung University$^{3}$}}
\\
{\tt\small {\{bocheng.ai11, jenjee, yctseng\}@nycu.edu.tw}}
\\
{\tt\small {fcchen@nhri.edu.tw}} \quad{\tt\small {azyen@cs.nycu.edu.tw}}
}

\begin{document}
\maketitle
\begin{abstract}
Large Language Models (LLMs) have recently been extended to the video domain, enabling sophisticated video-language understanding. However, existing Video LLMs often exhibit limitations in fine-grained temporal reasoning, restricting their ability to precisely attribute responses to specific video moments, especially under constrained supervision.
We introduce \textbf{DaMO}, a data-efficient Video LLM explicitly designed for accurate temporal reasoning and multimodal understanding. At its core, the proposed \textbf{Temporal-aware Fuseformer} employs a hierarchical dual-stream architecture that progressively captures temporal dynamics within each modality and effectively fuses complementary visual and audio information. To further enhance computational efficiency, DaMO integrates a global residual that reduces spatial redundancy while preserving essential semantic details.
We train DaMO via a structured four-stage progressive training paradigm, incrementally equipping the model with multimodal alignment, semantic grounding, and temporal reasoning capabilities. This work also contributes multiple datasets augmented from existing ones with LLM-generated temporally grounded QA pairs for tasks requiring temporal supervision.
Comprehensive experiments on temporal grounding and video QA benchmarks demonstrate that DaMO consistently surpasses prior methods, particularly in tasks demanding precise temporal alignment and reasoning. Our work establishes a promising direction for data-efficient video-language modeling.
Our project is released at \href{https://yue-723.github.io/DaMO-2025-page/}{https://yue-723.github.io/DaMO-2025-page/}
\end{abstract}
\begin{figure}[ht]
\centering
\includegraphics[width=0.85\columnwidth]{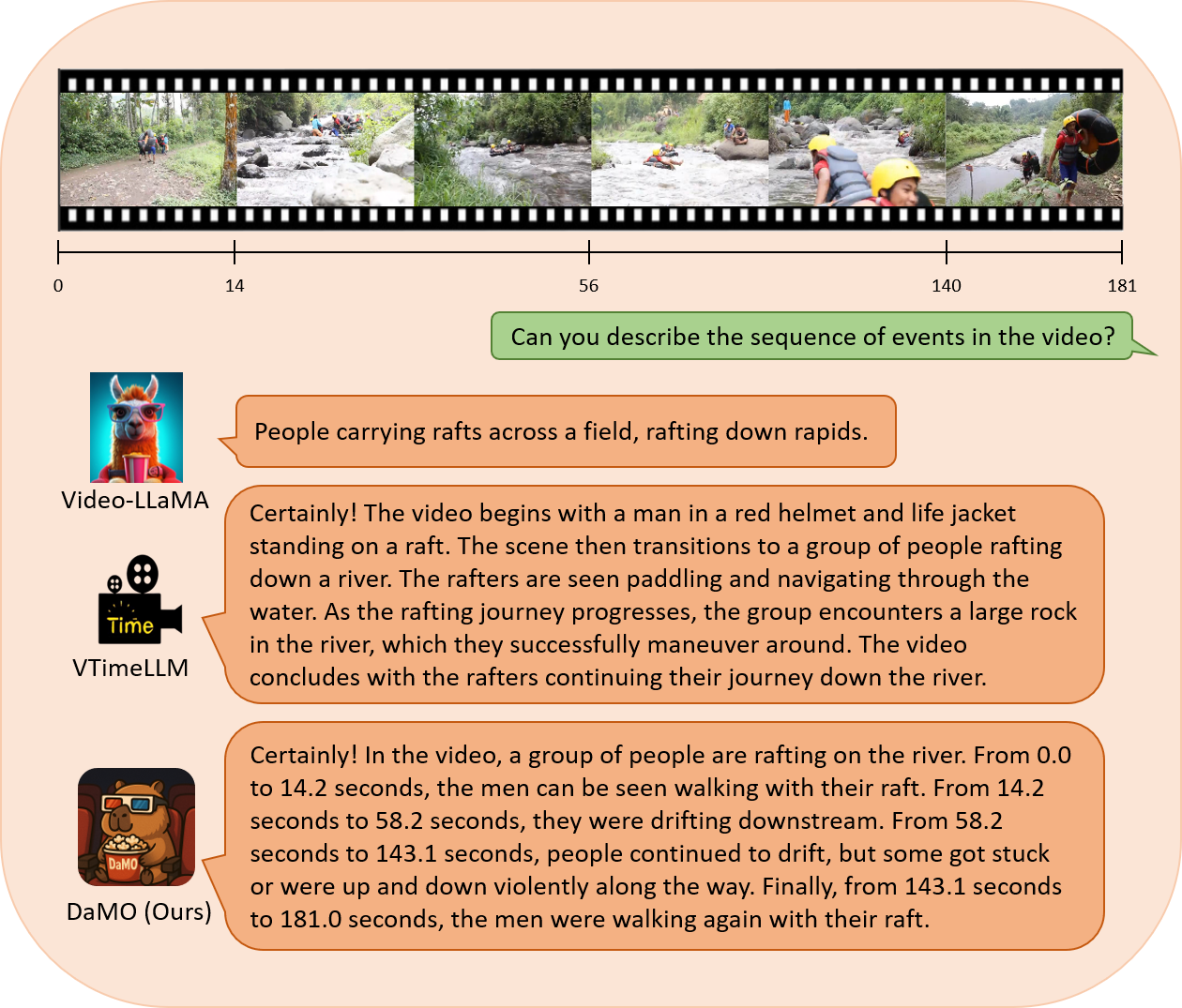}
\caption{Qualitative comparison on temporal reasoning in video-grounded QA. Given a temporal question grounded in a video clip, DaMO generates a more precise and temporally aligned response than Video-LLaMA~\cite{zhang2023video} and VTimeLLM~\cite{huang2024vtimellm}, showcasing superior temporal understanding.}
\label{fig:intro}
\end{figure}

\section{Introduction}
\label{sec:intro}
Recent advances in large language models (LLMs) have expanded their capabilities beyond text understanding~\cite{brown2020languagemodelsfewshotlearners,openai2023chatgpt,touvron2023llama,vicuna2023}, giving rise to multimodal LLMs (MLLMs) that are specialized for complex video-language understanding tasks~\cite{liu2024visual,lin2023video,luo2023valley,li2023mvbench}. These Video LLMs have demonstrated promising capabilities in applications such as video-grounded question answering, multimodal dialogues, and temporal event localization~\cite{huang2024vtimellm,qian2024momentor,qu2024chatvtg}. 

Despite notable progress, existing Video LLMs still face some limitations, particularly in effective multimodal integration, high data requirements, and limited temporal reasoning capability.
First, most current approaches primarily rely on visual modality alone~\cite{li2024videovista,zhang2024llamaadapter,xu2024pllava}, without fully leveraging the complementary modalities such as audio and textual transcripts within videos. While some recent works~\cite{zhang2023video,su2023pandagpt,lyu2023macawllm} attempt to incorporate audio or subtitle information, they usually treat each modality independently and let LLM perform cross-modality fusion, thus heavily relying on the inherent reasoning capability of LLMs to infer cross-modal relationships in an implicit manner. Such isolated processing hinders precise alignment of the temporal correlations between multimodal information.
Second, training powerful Video LLMs typically depends on large-scale datasets~\cite{zhang2023video, chen2023VideoLLM,li2023mvbench}, imposing substantial computation and storage costs. Such requirements significantly raise entry barriers and hinder rapid experimentation. 
Third, conventional approaches for spatial dimension reduction tend to discard critical global context information~\cite{li2024llamavid,li2023videochat,maaz2023video}, negatively impacting the quality of temporal feature extraction and fusion.

To address these challenges, we introduce DaMO, a data-efficient Video LLM explicitly designed for fine-grained temporal reasoning and multimodal integration. DaMO is empowered by several innovative architecture designs. At its core, it features {\em Temporal-aware Fuseformer (T-Fuseformer)}, a hierarchical dual-stream Transformer that progressively captures temporal dependencies within each modality and dynamically integrates complementary visual and audio information.
To further enhance efficiency without sacrificing representation quality, we propose a global residual that separately processes local contexts via pooling and global contexts through a lightweight feed-forward network, effectively preserving high-level semantics.

Complementing these architectural innovations, we adopt a structured four-stage progressive training paradigm to incrementally equip DaMO with multimodal alignment, semantic bridging, temporal perception, and dialogue reasoning capabilities:
\begin{enumerate}
\item \textbf{Video-Text Alignment}: 
It aligns temporally fused multimodal features with textual descriptions, establishing basic cross-modal grounding.
\item \textbf{Representation Bridging}: 
It projects fused features into the LLM-compatible semantic space.
\item \textbf{Temporal Perception Learning}: 
It explicitly teaches event localization and temporal relationships.
\item \textbf{Dialogue Tuning}: 
It fine-tunes the model on multi-turn dialogue to enhance conversational temporal reasoning.
\end{enumerate}
% \vspace{0.08in}
To make these trainings possible, this work also contributes multiple temporal QA datasets by enriching existing datasets through prompting with LLM, %GPT-based prompting, 
serving as a valuable asset for temporal reasoning benchmarking.

Extensive experiments on temporal grounding and video dialogue benchmarks confirm DaMO's strong performance, especially in tasks requiring fine-grained temporal reasoning under limited data and compute. As shown in Fig.~\ref{fig:intro}, DaMO produces more temporally aligned and coherent responses than existing Video LLMs, highlighting the effectiveness of our fusion architecture and progressive training strategy.

Our main contributions are summarized as follows:
\begin{itemize}
\item 
We propose DaMO, a data-efficient Video LLM designed for explicit multimodal temporal reasoning.
\item 
We introduce T-Fuseformer, a hierarchical dual-stream architecture for progressive temporal modeling and multimodal fusion.
\item 
To preserve global context while significantly reducing computational cost, we devise a global residual. % We devise a global residual to preserve global context while significantly reducing computational cost.
\item 
We present a four-stage progressive training paradigm to systematically enhance DaMO's multimodal alignment and temporal reasoning capabilities.
The training is facilitated by several newly curated datasets with LLM-generated temporal QA pairs.
\end{itemize}
\begin{figure*}[htbp]
  \centering
  \includegraphics[width=0.85\textwidth]{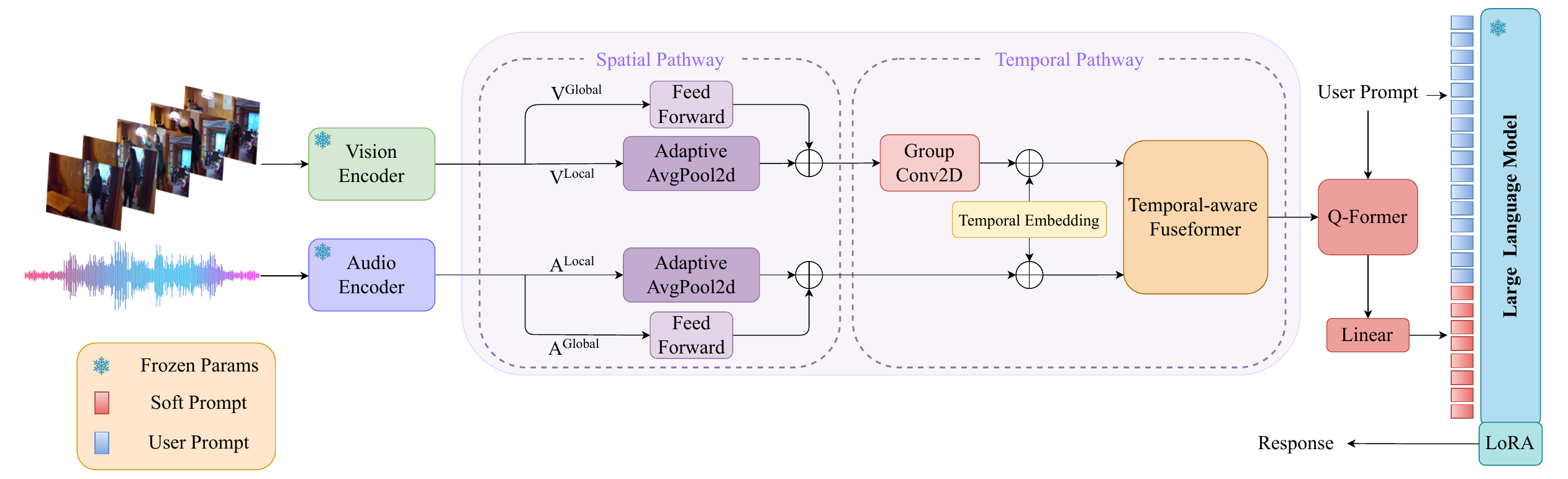}
  \caption{Overview of DaMO. Visual and audio features extracted by pretrained encoders undergo dimensionality reduction via a global residual, with grouped convolutions further compressing visual features along the temporal dimension. Before multimodal fusion, Temporal Embeddings are explicitly added to the modality-specific features. The Temporal-aware Fuseformer is designed to explicitly refine and integrate multimodal temporal representations, which are then projected into the embedding space of LLM adapted by LoRA via the Q-Former. The LLM is prompted by the concatenation of these embeddings and the user query for temporal reasoning.}
  \label{fig:overview}
\end{figure*}

\section{Related Works}
\label{sec:related}
\subsection{Video-Language Models}

Vision-language pretraining has considerably advanced multimodal representation learning.
CLIP~\cite{radford2021clip} established the first effective alignment between visual and textual modalities. 
Extending these insights to the video domain, pretraining frameworks such as Clip4Clip~\cite{Luo2021CLIP4Clip}, UMT~\cite{li2023unmasked}, and InternVideo2~\cite{wang2024internvideo2} have successfully leveraged temporal information. Additionally, VindLU~\cite{cheng2022vindlu} provided systematic analyses on critical aspects of video-language understanding, including temporal modeling and dataset design. However, these approaches typically depend on large-scale pretraining data, thus limiting their accessibility and scalability.

To more effectively integrate vision features into LLMs, BLIP-2~\cite{li2023blip} introduced Q-Former, which serves as an efficient bridge between pretrained vision encoders and frozen LLMs. Similarly, LLaVA~\cite{liu2024visual} demonstrated that lightweight instruction tuning can effectively adapt LLMs for visual tasks without extensive retraining. Recent video-centric models, including Video-LLaMA~\cite{zhang2023video}, VideoChatGPT~\cite{maaz2023video}, and VideoChat2~\cite{li2023mvbench}, extend these alignment strategies to video understanding, typically relying on simple frame-level pooling and extensive instruction-tuning datasets.

PLLaVA~\cite{xu2024pllava} introduced temporal pooling over video frames, smoothing feature distributions before inputting them into a LLM. While this approach enhances computational efficiency and representation stability, it inherently reduces fine-grained temporal details that are critical for tasks requiring precise temporal reasoning. 
Moreover, existing Video LLMs often treat complementary modalities, such as audio and subtitles, independently and separately, thereby significantly underutilizing cross-modal temporal complementarities and limiting their potential for richer multimodal reasoning~\cite{huang2021makesmultimodallearningbetter,xu2023multimodallearningtransformerssurvey}.

\subsection{Temporal Reasoning with Video LLMs}

Despite substantial progress in multimodal alignment, the ability to effectively reason about temporal dynamics remains a critical yet challenging aspect of video-language modeling. Temporal reasoning entails not only identifying and localizing specific events within videos but also comprehending and inferring their sequential and causal relationships. For example, given a video sequence in which an individual enters a room and subsequently sits, a temporally capable model should accurately respond to queries such as ``When does the person sit?'' or ``How long after entering does the person sit?''

Recent approaches have started addressing these challenges explicitly.
VTimeLLM~\cite{huang2024vtimellm} introduces a boundary-aware architecture designed explicitly for fine-grained temporal localization, along with a structured training pipeline and a temporally annotated dataset to support temporal grounding tasks. Similarly, Momentor~\cite{qian2024momentor} and ChatVTG~\cite{qu2024chatvtg} propose specialized training objectives targeting precise moment retrieval and temporal question answering, underscoring the importance of explicitly grounding model outputs to temporal segments.

However, existing methods still face significant limitations, particularly in modeling multi-scale temporal dependencies and comprehensively integrating temporally aligned multimodal information. These shortcomings are especially pronounced under constrained data or compute resources, highlighting the need for architectural innovations capable of preserving fine-grained temporal structures, efficiently leveraging multimodal signals, and reducing dependence on large-scale pretraining resources.

\section{Multimodal Orchestrator DaMO}

% DaMO is designed specifically for data-efficient training and effective temporal reasoning for Video LLM. 
% At its core, DaMO features multimodal T-Fuseformer, a dual-stream transformer architecture for robust temporal modeling and multimodal fusion combined with an efficient global residual strategy for spatial dimension reduction. 
% Furthermore, DaMO incorporates a four-stage progressive training strategy, achieving state-of-the-art results with significantly fewer training resources.

% \subsection{Model Overview}

Fig.~\ref{fig:overview} illustrates the architecture of DaMO, which is designed to perform multimodal temporal reasoning over audio-visual inputs.
The model takes as input an audio-visual stream along with a user prompt, and outputs a response generated by a pretrained language model. % The input is an audio-visual stream and a user prompt. A pretrained LLM is incorporated to provide a response.
To enable the LLM to reason about the audio-visual stream, the audio-visual stream is decomposed into two separate modalities: a silent video stream and an audio stream. %we first separate it into a silent video (or simply video) stream and an audio stream. 
Two pretrained visual and audio encoders are employed to separately extract their modality-specific features. 
These extracted features undergo a spatial path and a temporal path, and then are processed by Q-Former to project multimodal features into tokens that can adapt to the embedding space of the pretrained LLM accompanied by LoRA.
% At the end of this section, we will introduce our 4-stage training strategy.

\subsection{Pretrained Video and Audio Encoders}

DaMO employs the pretrained ViT-L/14 from EVA-CLIP~\cite{dosovitskiy2021vit,eva,EVA-CLIP} as its video encoder, and the pretrained Whisper-small~\cite{radford2022whisper} as its audio encoder. 
From the input audio-visual stream of any length, we uniformly sample $N$ image frames from it, denoted by $V \in \mathbb{R}^{N \times H \times W \times C}$, where $H$, $W$, and $C$ are the height, width, and channels of each frame.
We also uniformly sample the whole audio stream, putting them into $M$ length-$S$ segments and resulting in $A \in \mathbb{R}^{M \times S \times R}$, where $R$ is the audio frequency rate.

Each frame $V_i$ $(i=1,\dots,N)$ and audio segment $A_j$ $(j=1,\dots,M)$ is independently encoded by its corresponding encoder. In our implementation, each visual frame is resized to $336 \times 336$, and each audio segment length $S=30$ seconds with $R=16$ KHz. The resulting visual and audio features are denoted respectively as:
\begin{align}
\tilde{V}_i & = \text{ViT}(V_i)
    \in \mathbb{R}^{L_v \times D_v}
\\
\tilde{A}_j & = \text{Whisper}(A_j)
    \in \mathbb{R}^{L_a \times D_a}
\end{align}
where $L_v$ and $L_a$ denote the numbers of visual and audio tokens, and $D_v$ and $D_a$ are the corresponding feature dimensions, respectively.

\subsection{Spatial Pathway}

The high-dimensional features $[\tilde{V}_i]_{1:N}$ and $[\tilde{A}_j]_{1:M}$ extracted from pretrained encoders inherently contain extensive spatial redundancy, which may significantly increases computational load and complicate temporal reasoning. 
To reduce the computational complexity in the subsequent temporal pathway and multimodal fusion, we introduce a novel \textit{spatial average pooling with global residual} that is applied explicitly to the spatial dimension of the above features. 
Typically, aggressive spatial dimension reduction through conventional pooling methods leads to substantial loss in global contextual information, negatively impacting the representational quality. To address this limitation, our method explicitly separates features into \textit{local} and \textit{global} spatial components, thereby preserving critical global spatial semantics.

\begin{figure*}[htbp]
  \centering
  \includegraphics[width=0.8\textwidth]{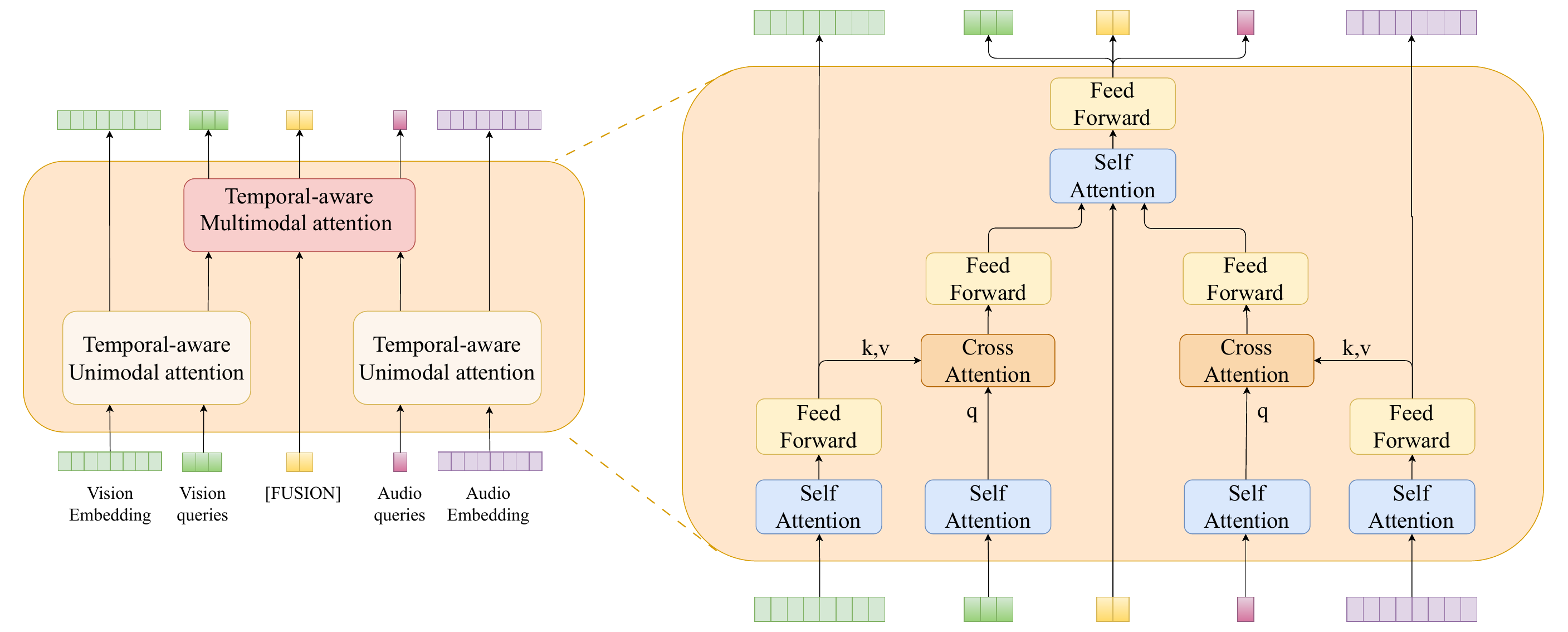}
  \caption{Architecture of T-Fuseformer. Each layer consists of unimodal attention and multimodal attention. Unimodal features are first refined via self-attention and then compressed via cross-attention with learnable queries. \texttt{FUSION} queries are introduced to attend to the compressed visual and audio features through self-attention and integrate multimodal information. Stacked layers progressively enhance temporal and cross-modal representations. The final \texttt{FUSION} queries serve as the temporally grounded representation to LLM.}
  \label{fig:fuseformer}
\end{figure*}

From the input feature maps
$[\tilde{V}_i]_{1:N}$, we obtain the global spatial feature $V^{\text{Global}}$ by extracting the $cls$ tokens, $\tilde{V}^{cls}_i$, for the visual modality.
However, for the audio modality, from $[\tilde{A}_j]_{1:M}$, we obtain global $A^{\text{Global}}$ by performing mean pooling over the spatial dimension $L_a$ as Whisper does not offer a $cls$ token.
\begin{align}
V^{\text{Global}} & =
[\tilde{V}^{cls}_i]_{1:N}
\in \mathbb{R}^{N \times D_v}
\\ 
A^{\text{Global}} & =
[ \text{mean} (\tilde{A}_j, \text{dim} = L_a)]_{1:M}
\in \mathbb{R}^{M \times D_a}
\end{align}

For the local spatial features, we derive $V^{\text{Local}} \in \mathbb{R}^{N \times (L_v - 1) \times D_v}$ by including all tokens except the $cls$ token from each frame and set $A^{\text{Local}} = [\tilde{A}_j]_{1:M}$ unchanged.

To effectively compress spatial redundancy, we apply adaptive average pooling to $V^{\text{Local}}$ and $A^{\text{Local}}$ separately, reducing both their lengths from $L_v -1$ and $L_a$ to the same smaller $L^{'}$. Simultaneously, the global features $V^{\text{Global}}$ and $A^{\text{Global}}$ are each refined through a lightweight feed-forward network (FFN). Finally, we fuse the compressed local representations with the refined global contexts through a residual addition:
\begin{equation}
X^{\text{Res}} = \text{AdaptiveAvgPool}(X^{\text{Local}}) + \text{FFN}(X^{\text{Global}})
\end{equation}
Here, $X = V$ or $A$, 
$V^{\text{Res}} \in \mathbb{R}^{N \times L^{'} \times D_v}$, and
$A^{\text{Res}} \in \mathbb{R}^{M \times L^{'} \times D_a}$.

By explicitly preserving global spatial context, the global residual alleviates semantic information loss typically encountered in standard spatial pooling approaches, while simultaneously reducing the computational complexity in downstream temporal modeling and multimodal fusion.

\subsection{Temporal Pathway}

The main function of this pathway is to perform temporal modeling and temporal data fusion. However, before this, the visual features still contain lots of temporally redundant information due to the uniform frame sampling step. To mitigate this, we additionally employ 2D grouped convolutions to perform dimensionality reduction along the temporal dimension of $V^{\text{Res}}$. This processing effectively reduces computational overhead while enabling DaMO to efficiently utilize richer temporal contexts.

Additionally, we explicitly incorporate temporal cues by adding Temporal Embeddings along the temporal dimension of each modality. Specifically, our Temporal Embedding is a combination of learnable positional embeddings and fixed sinusoidal embeddings, enabling DaMO to capture both flexible and structured temporal relationships effectively.

Subsequently, these compressed visual and audio features go through the proposed T-Fuseformer, a dual-stream transformer-based architecture explicitly designed for effective refinement and fusion of multimodal temporal information across multiple temporal scales. 
The architecture of T-Fuseformer is shown in Fig.~\ref{fig:fuseformer}. It consists of a stack of layers, each containing a hierarchy of (i) Temporal-aware Unimodal Attention and (ii) Temporal-aware Multimodal Attention. This architecture explicitly refines temporal representations within each modality and subsequently integrates complementary cross-modal information at multiple temporal scales, achieving effective multimodal fusion.
The inputs include 
(i) the visual embedding $GroupConv2d(V^{\text{Res}})$, 
(ii) the audio embedding $A^{\text{Res}}$, 
(iii) learnable visual queries, 
(iv) learnable audio queries, and 
(v) learnable \texttt{FUSION} queries.
At the end, only the \texttt{FUSION} queries are sent to the Q-Former.
The main purpose is to align the resulting multimodal representations with the semantic space of the frozen LLM via the Q-Former module of InstructBLIP~\cite{instructblip}.

\paragraph{Temporal-aware Unimodal Attention.} In this stage, we independently refine the temporal representations of each modality using modality-specific self-attention and feed-forward network (FFN). These result in temporally refined unimodal features. To compress the number of tokens and distill temporal information, we introduce a small number of learnable queries for each modality, which selectively aggregate the refined features through cross-attention.
These queries effectively attain three key objectives: (i)~removing redundant or uninformative temporal tokens, (ii)~retaining semantically salient temporal moments, and (iii)~producing a compact yet expressive representation. 
% Notably, even low-dimensional learnable queries can accurately capture the semantics of complex temporal sequences.

Empirically, we observe that the visual modality generally contains more detailed semantic information compared to audio~\cite{zhu2024visionxsurveymultimodallearning, aytar2017seehearreaddeep, nagrani2022attentionbottlenecksmultimodalfusion,cvpr2023PMR}.
% due to the inherent larger volume of visual information. 
Therefore, the ratio of the number learnable visual queries to the number of learnable audio queries is preset to a ratio $\alpha > 1$ (we set $\alpha = 3$ in our experiments). This ensures that both modalities are proportionally compressed without compromising representational quality. By stacking multiple such layers, our model progressively refines temporal representations at different granularity levels, enabling hierarchical temporal understanding and robust feature compression.

\paragraph{Temporal-aware multimodal Attention.} 
In this stage, we perform multimodal fusion using a lightweight cross-modal attention mechanism. Specifically, we introduce a small number of learnable \texttt{FUSION} queries, which are concatenated with the compressed temporal features from both visual and audio modalities. This combined sequence is then processed by a cross-modal self-attention layer followed by an FFN, facilitating effective multimodal interaction and integration.

These introduced \texttt{FUSION} queries dynamically aggregates complementary multimodal information by attending to the most salient and temporally aligned segments from both modalities. They are shared across all layers and progressively updated throughout the stacked T-Fuseformer layers, enabling them to increasingly capture temporally grounded multimodal representations.

\subsection{Token Projection and LoRA Adaptation}

The \texttt{FUSION} queries from the final T-Fuseformer layer thus serves as a compact, temporally grounded representation of the entire video, which is subsequently projected into the LLM's semantic space using the Q-Former, followed by a linear transformation. These projected multimodal embeddings serve as soft prompts, which are concatenated with user-generated textual queries and fed into the pretrained LLM.
To adapt to these \texttt{FUSION} queries, LoRA~\cite{hu2022lora} is applied to the frozen LLM %the frozen LLM is enhanced by LoRA~\cite{hu2022lora} 
to generate temporally grounded and contextually precise responses.

% Compared to conventional approaches relying on flat pooling or naive concatenation, our Temporal-aware Fuseformer achieves significantly enhanced temporal discriminability, multimodal interpretability, and computational efficiency through hierarchical refinement and effective modality-aware attention.

\subsection{Data-Efficient Progressive Training Paradigm}

To systematically equip DaMO with robust multimodal reasoning capabilities, we propose a structured four-stage progressive training strategy. This hierarchical approach incrementally enhances the model's abilities from basic multimodal alignment, through semantic bridging with the LLM's representation space, and ultimately toward sophisticated conversational audio-visual understanding. Through this structured paradigm, DaMO effectively acquires strong temporal reasoning skills, even when trained on limited data.

\paragraph{Stage 1: Video-Text Alignment.}
In this initial stage, our primary goal is to align temporally fused multimodal representations with their corresponding textual descriptions by leveraging the structure of Q-Former. Differing from conventional methods that individually align unimodal features, we directly align the multimodal audio-visual features obtained from the T-Fuseformer with the textual information of the input video. 
The fused audio-visual embeddings are projected into the textual semantic space via the Q-Former. 

During training, the linear projection layers and the frozen LLM are not involved.
We adopt three complementary training objectives widely employed in vision-language pretraining~\cite{nips2021albef,li2022blip,li2023blip,wang2022omnivl,cheng2022vindlu}: 
(i)~{\em Vision-Text Contrastive (VTC) learning} for instance-level alignment, 
(ii)~{\em Vision-Text Matching (VTM)} for pairwise correspondence discrimination, and 
(iii)~{\em Vision-grounded Text Generation (VTG)} to strengthen semantic grounding at the local context level.

This stage utilizes approximately 1.5M video-text pairs sampled from the InternVid-10M dataset~\cite{wang2024internvidlargescalevideotextdataset}, providing diverse and semantically rich supervision to build strong multimodal representations.

\paragraph{Stage 2: Representation Bridging.}
In this stage, our objective shifts toward bridging the semantic gap between temporally fused multimodal embeddings and the LLM's internal representation space. Here, the fused embeddings from the T-Fuseformer are processed by the Q-Former followed by a lightweight linear transformation and then fed into the frozen LLM. Importantly, the LLM parameters remain fully frozen, and LoRA adaptation is {\em not} applied at this stage.

We employ the VTG objective to condition the frozen LLM, guiding it to generate temporally coherent and contextually accurate responses. To facilitate this semantic alignment, we curate a training corpus comprising approximately 300K QA pairs. 
This dataset integrates several sources, including general video-based QA data from VideoInstruct-100K~\cite{maaz2023video} and a refined version of AVSD processed by Macaw-LLM~\cite{lyu2023macawllm}.
To enrich the temporal grounding component, we additionally incorporate four temporally annotated datasets: QVHighlight~\cite{lei2021qvhighlight}, Charades-STA~\cite{Charades-STA}, ActivityNet~\cite{krishna2017dense}, and 100K samples selected from Koala36M~\cite{wang2024koala36mlargescalevideodataset}. Building upon these resources, we leverage LLM-based prompting to regenerate approximately 150K QA pairs with explicit temporal grounding derived from their original segment annotations.
The augmentation details are provided in Supplementary.

By combining general QA datasets with the LLM-augmented datasets, we equip DaMO with diverse semantic understanding and fine-grained temporal alignment capabilities, providing robust representation bridging.

\paragraph{Stage 3: Temporal Perception Learning.}
The goal of this stage is to explicitly strengthen DaMO's ability to reason about temporally grounded multimodal information. While the previous stage emphasizes semantic coherence, we now specifically focus on training the model to attend to, reference, and reason over temporal contexts.

To efficiently adapt the frozen LLM for temporal reasoning tasks, we introduce Low-Rank Adaptation (LoRA) to fine-tune the LLM. The whole pipeline in Stage 2, together with the LoRA module, are trained.
We continue employing the VTG objective, now explicitly targeted toward supervising temporally-aware causal reasoning. This training guides DaMO to generate responses explicitly anchored to particular temporal segments in videos.
The same set of 300K QA pairs used in Stage 2 is leveraged, ensuring knowledge continuity and data efficiency. This stage is pivotal for transitioning DaMO's capabilities from semantic comprehension toward temporally grounded understanding.

\paragraph{Stage 4: Dialogue Tuning.} 
The objective in this stage is to adapt DaMO for temporally grounded multi-turn dialogue, enabling it to produce coherent responses while reasoning over temporally structured video features.
Building upon the training outcomes of Stage 3, we continue fine-tuning the full pipeline using the VTG objective.  This stage leverages a curated 39K dialogue-oriented dataset, comprising general dialogues from VideoChat2~\cite{li2023mvbench} and ActivityNet-based dialogues adapted from VTimeLLM~\cite{huang2024vtimellm}. Additionally, we augment  DSTC10-AVSD~\cite{DSTC10-AVSD} with LLM-based prompting to inject temporal annotations aligned with its dialogue flows, ensuring it to support training temporally grounded dialogue generation.
The details are provided in Supplementary.

To summarize, the progressive training paradigm enables DaMO to acquire advanced conversational capabilities, nuanced temporal reasoning, and the ability to generate contextually coherent and temporally grounded responses, all while maintaining strong data efficiency. In addition to architectural contributions, in Stages 2-4, we further augment several existing datasets by enriching them with temporally grounded information through LLM-based prompting. 
These augmented datasets are released together with the paper to facilitate future research.

\section{Experiment Results}
\subsection{Implementation Details}

To sample an input audio-visual stream, we set $N=24$ visual frames, $M=8$ audio segments, and $S=30$ seconds. 
To reduce computational complexity, the $GroupConv2d$ in the temporal pathway uses $N/3$ groups. 
We adopt the frozen LLaVA-v1.6-Mistral-7B~\cite{liu2024improved,jiang2023mistral7b} as our pretrained LLM by adapting it using LoRA with a rank of 32 and scaling factor $\alpha = 64$. Within T-Fuseformer, we utilize 192 learnable visual query tokens, 64 learnable audio query tokens, and 128 \texttt{FUSION} tokens across all layers, each query with a dimension of 768. 
All models are optimized using AdamW with learning rate $1e^{-4}$ and weight decay 0.02 on 4× NVIDIA A100 GPUs.

In our 4-stage progressive training, Stage 1 pretrains on 1.5 million video-text pairs with a batch size of 40 for 4 epochs, requiring about 3.5 days. 
Stages 2 and 3 share the same 300K QA pairs with a batch size of 8.
An epoch in stage 2 takes about 5 hours.
Stage 3 runs for 2 epochs, taking about 12 hours. 
Stage 4 runs on the dialogue-oriented dataset comprising 39K conversation examples for 2 epochs with the same batch size, taking about 2 hours.

\subsection{Zero-Shot Video Retrieval Benchmarks}
\begin{table*}[ht]
    \small
    \centering
    \resizebox{.72\linewidth}{!}{
        \begin{tabular}{l|c|ccc|ccc}
        \toprule
        \multirow{2}{*}{\textbf{Method}} & \multirow{2}{*}{\textbf{Training Data}} & \multicolumn{3}{c|}{\textbf{MSR-VTT}} & \multicolumn{3}{c}{\textbf{MSVD}} \\ \cline{3-8}
                                                               &         & R@1     & R@5     & R@10     & R@1     & R@5     & R@10  \\ \hline
        OmniVL~\cite{wang2022omnivl}                           & 14M     & \underline{34.6}    & 58.4    & 66.6     & -       & -       & -          \\ 
        UMT-L~\cite{li2023unmasked}                            & 5M      & 33.1    & 58.1    & 66.7     & 44.3    & 73.3    & 82.4      \\ 
        CLIP4Clip$^\dagger$~\cite{Luo2021CLIP4Clip}            & 400M    & 32.0    & 57.0    & 66.9     & 38.5    & 76.8    & -          \\ 
        InternVideo2-6B$^\dagger$~\cite{wang2024internvideo2}  & 404M    & \textbf{55.9} & \textbf{78.3} & \textbf{85.1} & \underline{59.3} & \underline{84.4} & \textbf{89.6} \\
        \hline
        DaMO (Ours)                                              & 1.5M    & \underline{34.6}    & \underline{58.7}    & \underline{67.2}     & \textbf{64.8}    & \textbf{85.2}    & \underline{89.5} \\
        \bottomrule
        \end{tabular}}
        \caption{Zero-shot video retrieval benchmarks on MSR-VTT and MSVD (Recall@1/5/10). Models marked with $^\dagger$ indicate pretraining on large-scale video-text data. (boldface=best, underline=runner-up)}
        \label{tab:zs-vr}
\end{table*}
We first evaluate DaMO's capability to extract discriminative multimodal representations.
This evaluation specifically assesses the effectiveness of Stage 1 training, focusing on DaMO's multimodal temporal alignment ability without any retrieval-specific fine-tuning. 
We test the zero-shot video retrieval task on MSR-VTT~\cite{MSRVTT} and MSVD~\cite{MSVD}. 
Following standard protocols, we report Recall@1, Recall@5, and Recall@10 metrics.

We compare DaMO with several strong baselines, including OmniVL~\cite{wang2022omnivl}, UMT~\cite{li2023unmasked}, CLIP4Clip~\cite{Luo2021CLIP4Clip}, and InternVideo2-6B~\cite{wang2024internvideo2}. Notably, CLIP4Clip and InternVideo2-6B leverage significantly larger training corpora.

As shown in Table~\ref{tab:zs-vr}, DaMO achieves competitive results despite training on considerably less data. On MSR-VTT, DaMO ranks second only to InternVideo2-6B, while on MSVD, it surpasses InternVideo2-6B at Recall@1 and Recall@5, highlighting its strong generalization ability and data efficiency. These results validate the effectiveness of our temporal fusion architecture and multimodal alignment strategy under resource-constrained conditions.

\begin{table*}[hb]
\centering
\resizebox{0.7\linewidth}{!}{
    \begin{tabular}{l|cccc|cccc}
        \toprule
        \multirow{2}{*}{Method} & \multicolumn{4}{c|}{Charades-STA} & \multicolumn{4}{c}{ActivityNet-Captions} \\ \cline{2-9}
                                                 & R@0.3 & R@0.5 & R@0.7 & mIoU      & R@0.3 & R@0.5 & R@0.7 & mIoU \\\hline
            VideoLLaMA~\cite{zhang2023video}     & 10.4  & 3.8   & 0.9   & 7.1       & 6.9   & 2.1   & 0.8   & 6.5\\
            VideoChat~\cite{li2023videochat}     & 9.0   & 3.3   & 1.3   & 6.5       & 8.8   & 3.7   & 1.5   & 7.2 \\
            VideoChatGPT~\cite{maaz2023video}    & 20.0  & 7.7   & 1.7   & 13.7      & 26.4  & 13.6  & 6.1   & 18.9\\
            VTimeLLM~\cite{huang2024vtimellm}    & 51.0  & 27.5  & 11.4  & 31.2      & 44.0  & 27.8  & \underline{14.3}  & 30.4\\
            VTimeLLM-13B~\cite{huang2024vtimellm} & \textbf{55.3}   & \underline{34.3}  & 14.7  & 34.6     & \underline{44.8}  & \underline{29.5}  & 14.2  & \underline{31.4}\\
            Momentor~\cite{qian2024momentor}     & 42.6  & 26.6  & 11.6  & 28.5      & 42.9  & 23.0  & 12.4  & 29.3\\
            ChatVTG~\cite{qu2024chatvtg}         & \underline{52.7}  & 33.0  & \underline{15.9}  & \textbf{34.9}      & 40.7  & 22.5  & 9.4   &27.2\\
        \hline
            DaMO (Ours)                            & 50.1  & \textbf{35.5}  & \textbf{21.2}  & \underline{34.8}     & \textbf{57.0}  & \textbf{39.7}  & \textbf{23.9}  & \textbf{40.3}\\
        \bottomrule
    \end{tabular}}
    \caption{Video-LLM temporal grounding benchmarks on Charades-STA and ActivityNet-Captions. 
    % DaMO achieves state-of-the-art performance, demonstrating superior capabilities in localizing and reasoning over precise temporal video segments. 
    All model sizes are 7B unless otherwise specified.}
    \label{tab:main}
\end{table*}

\subsection{Video-LLM Temporal Grounding Benchmarks}

To evaluate fine-grained temporal localization capabilities, we conduct experiments on two standard temporal grounding benchmarks: Charades-STA~\cite{Charades-STA} and ActivityNet-Captions~\cite{krishna2017dense}. To ensure reliable evaluation, we perform an additional instruction-following fine-tuning step on top of the pretrained DaMO. Specifically, we freeze all DaMO parameters and train a lightweight LoRA adapter (rank reduced by half) to generate structured temporal segmentation predictions with the output format: {\small\texttt{"There are X relevant segments: [[start$_1$, end$_1$], [start$_2$, end$_2$], ...]"}}. This explicit formatting enables numerical comparison.

As shown in Table~\ref{tab:main}, DaMO achieves strong overall performance across both benchmarks, significantly outperforming existing Video LLM methods. These outcomes affirm our approach's superior temporal localization performance, validating the effectiveness of our temporal fusion mechanism and progressive training paradigm.

\subsection{Temporal Dialogue Understanding}

To assess DaMO's capability in conducting temporally grounded dialogues, we perform experiments using VCGbench~\cite{maaz2023video}, a benchmark designed to assess generative video-language models across five critical dimensions: Correctness of Information, Detail Orientation, Contextual Understanding, Temporal Understanding, and Consistency.
VCGbench comprises videos from the densely annotated ActivityNet-200 dataset~\cite{caba2015activitynet}, with GPT-3.5-based automatic evaluation scores from 1 to 5 across these dimensions.

As shown in Table~\ref{tab:VCG_Bench}, DaMO achieves the highest score in the Temporal Understanding metric, clearly highlighting its capability to accurately localize and reason about temporally grounded content. 
Its performance across other metrics remains moderate, likely due to the instruction-tuning phase primarily focusing on temporally grounded QA, thus limiting broader conversational exposure.

\begin{table*}[ht]
    \centering
    \resizebox{.7\linewidth}{!}{
    \begin{tabular}{lccccc}
        \toprule
        Method & \thead{Correctness of\\Information} &  \thead{Detail\\Orientation} & \thead{Contextual\\Understanding} & \thead{Temporal\\Understanding} & \thead{Consistency} \\
        \midrule
        VideoLLaMA~\cite{zhang2023video}        &1.96               &2.18               &2.16       &1.82       &1.79 \\
        VideoChat~\cite{li2023videochat}        & 2.23              & 2.50              & 2.53      & 1.94      & 2.24 \\
        Video-ChatGPT~\cite{maaz2023video}      & 2.50              & 2.57              & 2.69      & 2.16      & 2.20 \\
        BT-Adapter~\cite{cvpr2024BT-Adapter}    & 2.68              & 2.69              & 3.27      & 2.34      & 2.46 \\
        MovieChat~\cite{song2023moviechat}      & 2.76              & 2.93              & 3.01      & 2.24      & 2.42 \\
        Chat-UniVi~\cite{jin2023chatunivi}      & 2.89              & 2.91              & 3.46      & 2.89      &\underline{2.81} \\
        LLaMA-VID~\cite{li2024llamavid}         &2.96               &3.00               &3.53 &2.46 &2.51 \\
        Vista-LLaMA~\cite{VISTA-LLAMA}          & 2.44              & 2.64              & 3.18 & 2.26 & 2.31 \\
        VTimeLLM~\cite{huang2024vtimellm}       & 2.78              & \textbf{3.10}     & 3.40 & 2.49 & 2.47 \\
        VideoChat2~\cite{li2023mvbench}         & 3.02              & 2.88              & 3.51 & 2.66 & \underline{2.81} \\
        ST-LLM\cite{eccv2024stllm}              & \textbf{3.23}     & \underline{3.05}  & \textbf{3.74} & \underline{2.93} & \underline{2.81} \\
        PLLaVA\cite{xu2024pllava}               & \underline{3.21}  & 2.86              & \underline{3.62} & 2.33 & \textbf{2.93} \\
        \midrule
        DaMO (Ours)                               & 2.89 & 2.55 & 3.21 & \textbf{3.10} & 2.53 \\
        \bottomrule
    \end{tabular}}
    \caption{Video dialogue benchmarks on VCGbench. DaMO notably outperforms other models in Temporal Understanding.}
    \label{tab:VCG_Bench}
\end{table*}

\subsection{Ablation Study}
\paragraph{LoRA Configuration.} 
We vary the LoRA rank in \{16, 32, 64\} for LLM. 
As shown in Table~\ref{tab:ablation1}, rank 32 yields optimal performance, especially in Temporal Understanding. Interestingly, increasing the LoRA rank further to 64 negatively impacts performance, suggesting that excessive adaptation capacity may cause overfitting or disrupt previously learned temporal alignment, particularly under limited data scenarios.

\begin{table}[h]
    \centering
    \resizebox{.95\linewidth}{!}{
    \begin{tabular}{lccccc}
        \toprule
        \thead[l]{LoRA\\configuration} & \thead{Correctness of\\Information} &  \thead{Detail\\Orientation} & \thead{Contextual\\Understanding} & \thead{Temporal\\Understanding} & \thead{Consistency} \\
        \midrule
        r=16,  $\alpha$=32 & \textbf{2.91} & 2.53 & \textbf{3.21} & 2.98 & 2.51 \\
        r=32,  $\alpha$=64 & 2.89 & \textbf{2.55} & \textbf{3.21} & \textbf{3.10} & \textbf{2.53} \\
        r=64,  $\alpha$=128 & 2.84 & 2.54 & 3.16 & 2.96 & 2.37 \\
        \bottomrule
    \end{tabular}}
    \caption{Ablation study of LoRA configuration on VCGbench.}
    \label{tab:ablation1}
\end{table}

\paragraph{Training Strategy.}
We further evaluate the effectiveness of our progressive training strategy by conducting ablations on VCGbench.
Several simplified variants are investigated. 
Specifically, after completing Stage 2, we compare: (1) skipping both Stages 3 and 4, (2) performing only Stage 3 (temporal reasoning), (3) performing only Stage 4 (dialogue tuning), and (4) jointly training Stages 3 and 4 (named JointS3S4).

\begin{table}[h]
\small
\centering
\resizebox{.95\linewidth}{!}{
    \begin{tabular}{lccccc}
        \toprule
        \thead{Training\\strategy} & \thead{Correctness of\\Information} &  \thead{Detail\\Orientation} & \thead{Contextual\\Understanding} & \thead{Temporal\\Understanding} & \thead{Consistency} \\
        \midrule
        w/o S3 and S4                   & 1.56 & 1.88 & 1.98 & 1.85 & 1.72 \\
        w/o S4                              & 2.84 & 2.54 & 3.15 & 2.41 & 2.35 \\
        w/o S3                              & 2.71 & 2.41 & 3.06 & 3.08 & 2.34 \\
        JointS3S4                                  & 2.83 & \textbf{2.55} & 3.19 & 2.65 & 2.35 \\
        DaMO                           & \textbf{2.89} & \textbf{2.55} & \textbf{3.21} & \textbf{3.10} & \textbf{2.53} \\
        \bottomrule
    \end{tabular}}
\caption{Ablation study of training strategy on VCGbench. (S=stage)}
\label{tab:ablation2}
\end{table}

As shown in Table~\ref{tab:ablation2}, our progressive strategy consistently achieves the best overall performance, particularly excelling in Temporal Understanding and Consistency. This analysis confirms that explicitly structuring the training into distinct 4 stages is crucial for effectively enhancing DaMO’s temporal reasoning and dialogue capabilities.

\section{Conclusions}

We have proposed DaMO, a data-efficient video-language model designed to perform fine-grained temporal reasoning and multimodal understanding. 
DaMO explicitly models temporal relationships and effectively leverages complementary semantic cues across visual and audio modalities, significantly achieving state-of-the-art performance in temporal reasoning.
DaMO incorporates Temporal-aware Fuseformer, a hierarchical multimodal fusion architecture that progressively refines temporal representations within each modality and dynamically integrates visual and audio information. To further improve computational efficiency without compromising representational quality, DaMO employs a global residual for spatial dimensionality reduction.
It is validated that the proposed four-stage progressive training paradigm substantially boosts data efficiency, reducing reliance on extensive training data.
Extensive experiments demonstrate DaMO’s strong performance in zero-shot video retrieval, temporal grounding, and dialogue tasks, validating its effectiveness in advancing data-efficient temporal reasoning.

Additionally, this work contributes several temporal QA datasets that are enhanced from existing video-language datasets via LLM-based prompting.
Further dataset information is available in Supplementary.
In the future, we plan to further enhance the semantic comprehension and reasoning capabilities of DaMO beyond temporal aspects and incorporate additional modalities.

{
    \small
    \bibliographystyle{ieeenat_fullname}
    \bibliography{main}
}

\clearpage
\setcounter{page}{1}
\maketitlesupplementary
\section{Additional Qualitative Examples}

To further showcase the capabilities of DaMO, we present two qualitative examples highlighting different aspects of its performance. These examples demonstrate DaMO’s proficiency in understanding complex visual narratives within multi-turn dialogues, as well as its effectiveness in precise temporal localization and reasoning during audio-visual interactions.

\begin{figure}[h]
  \centering
  \includegraphics[width=0.95\linewidth]{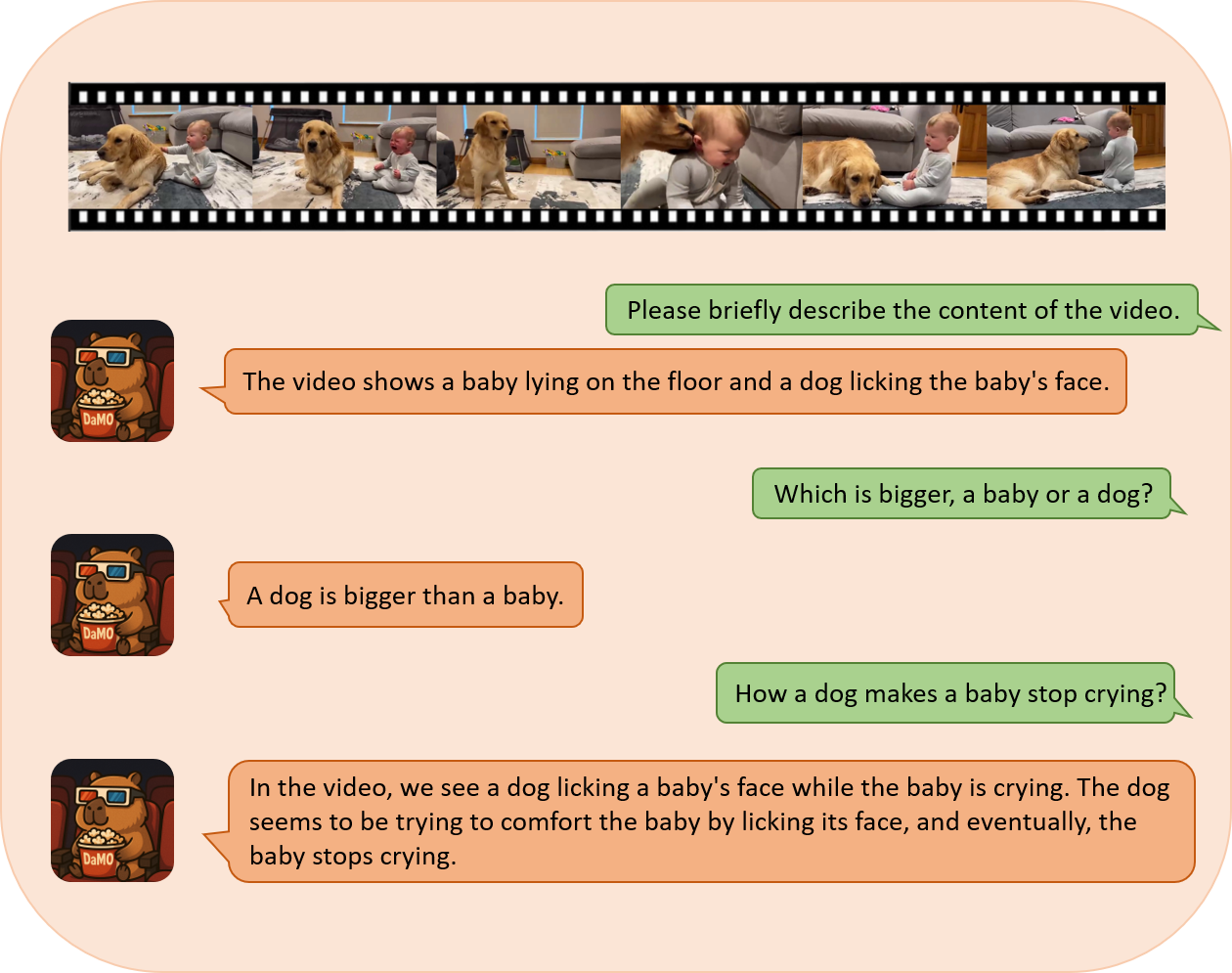}
  \caption{Multi-turn Visual Understanding in Dialogue.}
  \label{fig:sup1}
\end{figure}

\paragraph{Example 1: Multi-turn Visual Understanding.}
Fig.~\ref{fig:sup1} illustrates DaMO’s ability to comprehend detailed visual content through multi-turn dialogue. Given a series of user questions focusing on identifying differences between objects or interpreting scene narratives, DaMO consistently delivers accurate and contextually relevant responses. This demonstrates its coherent grasp of visual semantics and high-level event comprehension.

\begin{figure}[h]
  \centering
  \includegraphics[width=0.95\linewidth]{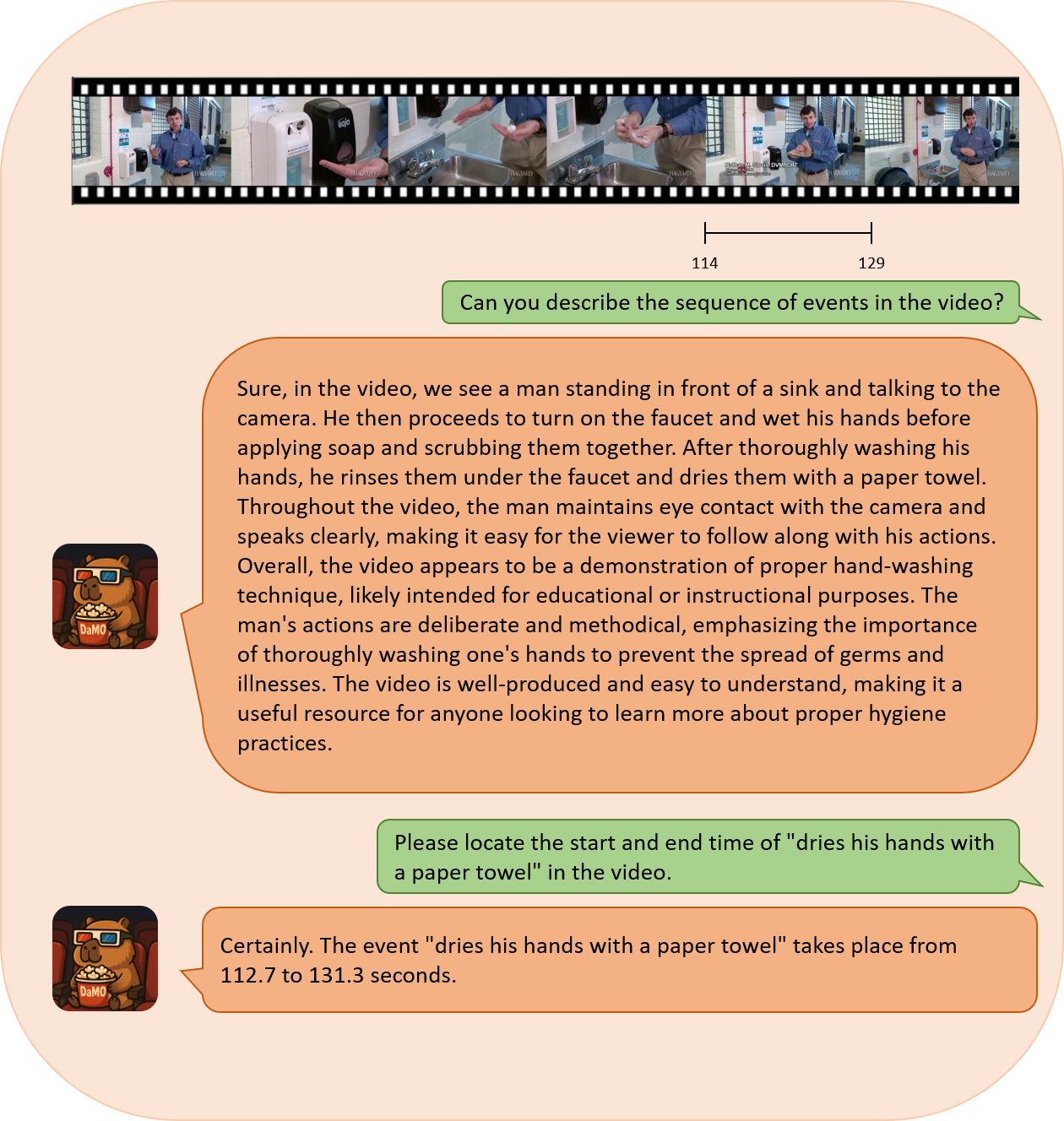}
  \caption{Temporal Localization and Reasoning.}
  \label{fig:sup2}
\end{figure}

\paragraph{Example 2: Temporal Localization and Reasoning.}
In the scenario of Fig.~\ref{fig:sup2}, DaMO responds to a user query requiring precise identification and localization of a specific event within the video. The model successfully anchors its response to the appropriate temporal segment and explicitly articulates its reasoning based on contextual and temporal cues. This example emphasizes DaMO’s capability in fine-grained temporal analysis and generating temporally precise dialogue responses.

\section{LLM-augmented Temporal QA Datasets}

To train DaMO to learn to reason about temporal information, we need to curate a temporal QA dataset without depending on costly large-scale manual annotations.
There exist several event-level temporally annotated datasets.
While these datasets contain valuable temporal signals, their original annotations are either limited to span-based descriptions without QA structure, or lack fine-grained temporal grounding within dialogues. This restricts their direct applicability to temporal QA tasks, especially for those requiring alignment between natural language queries and specific video moments.
Therefore, we construct a LLM-augmented dataset from the following datasets: 
(i) QVHighlight~\cite{lei2021qvhighlight}, 
(ii) Charades-STA~\cite{Charades-STA},
(iii) ActivityNet~\cite{krishna2017dense}, 
(iv) 100K selected samples from Koala36M~\cite{wang2024koala36mlargescalevideodataset}, 
and
(v) 9K multi-turn dialogues from DSTC10-AVSD~\cite{DSTC10-AVSD} enriched with explicit temporal references.
The curated dataset comprises 150K QA pairs and is used in DaMO's training stages 2 and 3.
The new benchmark is also released to the research community.

To augment these datasets, we leverage LLM-based prompting to systematically enhance their annotations by generating high-quality, temporally grounded QA pairs. Our contributions are twofold: (1) transforming timestamped event descriptions into natural QA format, and (2) integrating explicit temporal references into multi-turn dialogues. This augmentation provides more precise supervision for training temporal reasoning models.

\begin{figure*}
  \centering
  \includegraphics[width=0.95\linewidth]{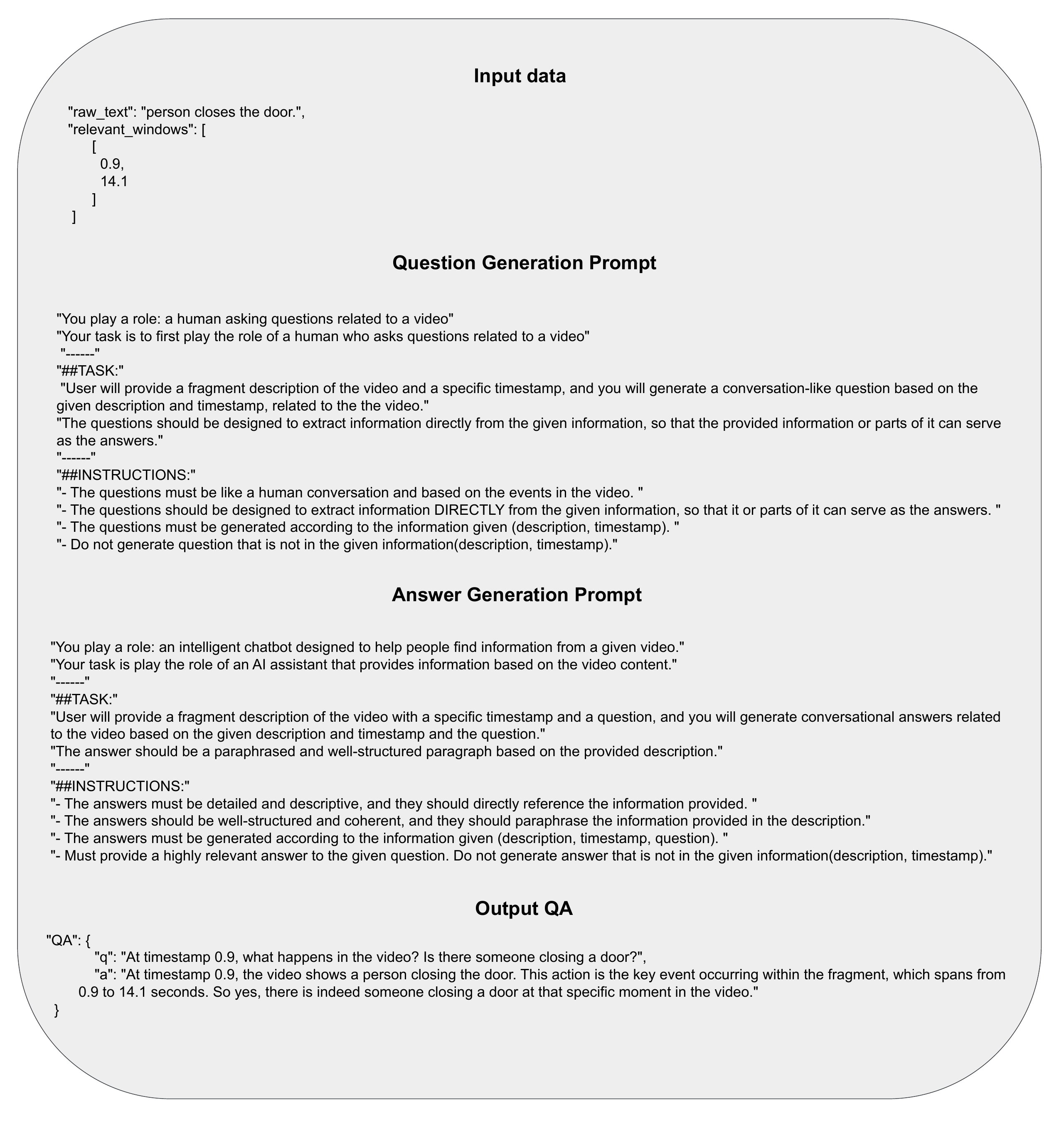}
  \caption{LLM-based QA generation from timestamped descriptions.}
  \label{fig:inst}
\end{figure*}

\paragraph{QA Generation from Timestamped Descriptions.}
For datasets with event-level temporal spans, such as Charades-STA~\cite{Charades-STA}, we design prompts that instruct LLM to convert timestamped segments into natural questions and corresponding answers.
An example is in Fig.~\ref{fig:inst}. This enables learning temporal grounding in a QA context, bridging the gap between annotation spans and linguistic query-answer structures.

\begin{figure*}
  \centering
  \includegraphics[width=0.95\linewidth]{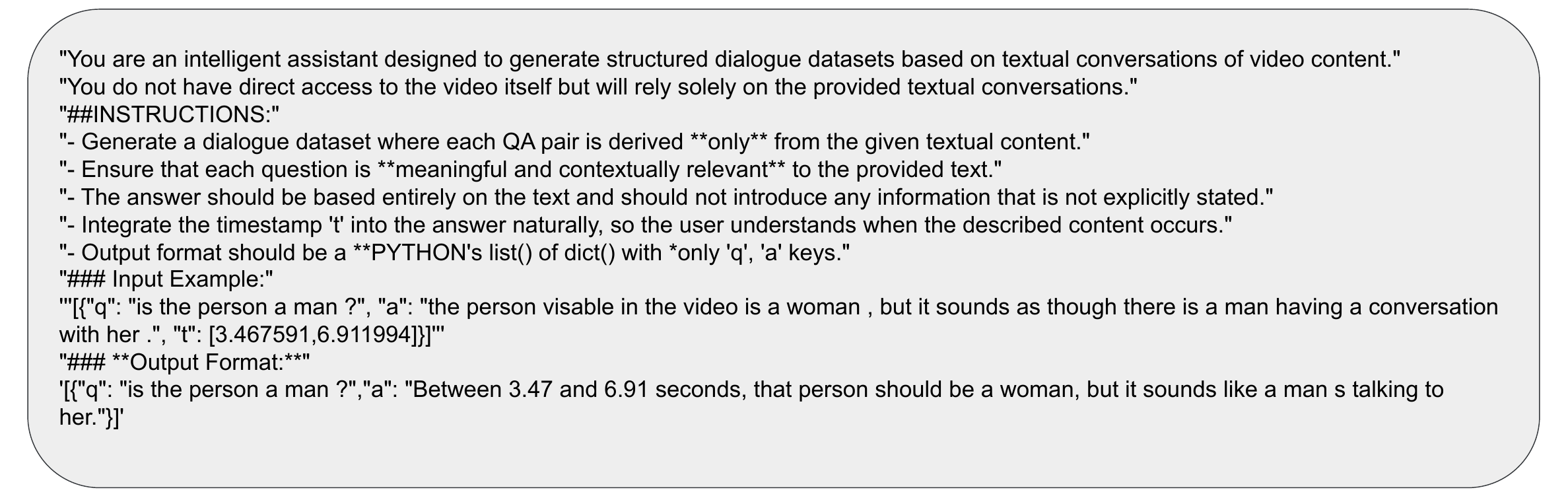}
  \caption{LLM-based temporal enrichment in multi-turn dialogue.}
  \label{fig:inst2}
\end{figure*}

\paragraph{Temporal Enrichment in Multi-turn Dialogue.}
For multi-turn dialogues with associated temporal segments (e.g., DSTC10-AVSD ~\cite{DSTC10-AVSD}), we use LLM to integrate these timestamps into the dialogue naturally.
An example is in Fig.~\ref{fig:inst2}. The prompt instructs LLM to revise or regenerate responses so that they include explicit temporal references tied to the original annotations, improving temporal traceability and alignment throughout the conversation.

\section{Evaluation Setup for Instruction Following}
To enable controlled and reproducible evaluation of temporal grounding, we apply an additional instruction-following fine-tuning step that standardizes the model's output format. This allows for direct, automated comparison with ground-truth annotations without requiring post-hoc heuristics or response parsing.

Specifically, we fine-tune a lightweight LoRA adapter with a rank set to less than half of the configuration used in our main training. This compact adapter is trained to condition the model to output temporally grounded predictions in a structured format that enumerates all relevant video segments in response to a natural language query.

\begin{figure*}
  \centering
  \includegraphics[width=0.95\linewidth]{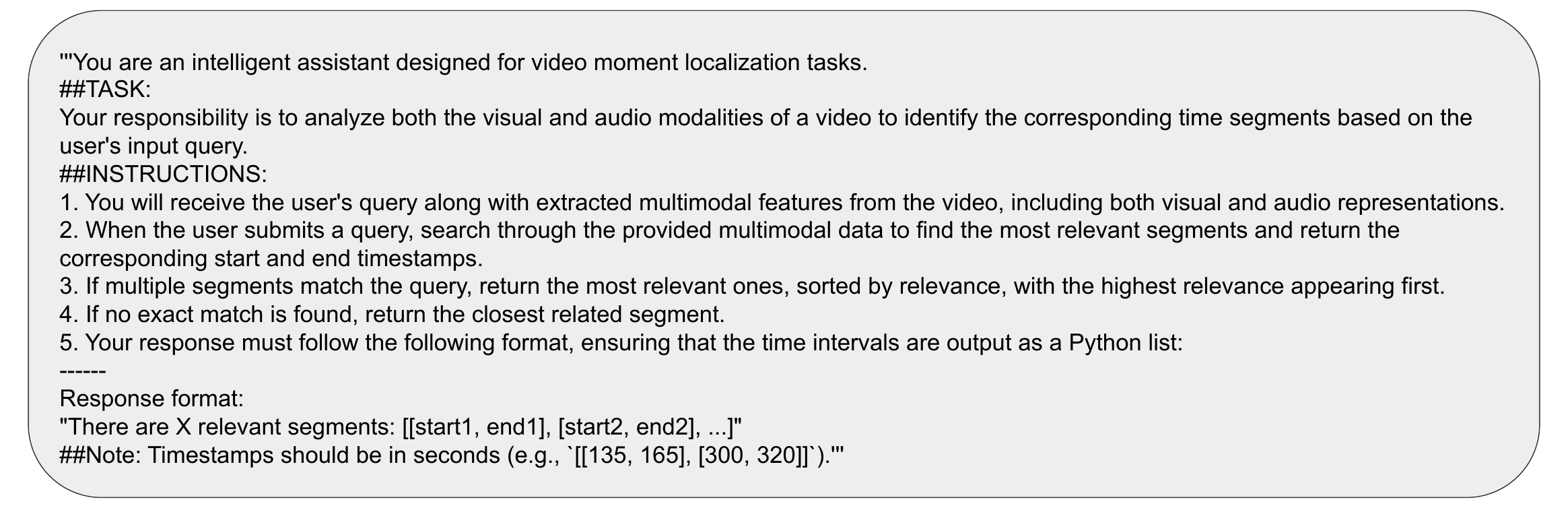}
  \caption{Instruction-following prompt for temporal grounding evaluation.}
  \label{fig:inst3}
\end{figure*}

To support this process, we design a simple and consistent prompt that instructs the model to return its prediction in a list-style format. This enables easy parsing and compatibility with standard evaluation metrics such as temporal Intersection-over-Union (tIoU). The instruction prompt used for this fine-tuning is shown in Fig.~\ref{fig:inst3}.

\end{document}